\documentclass[runningheads]{llncs}
\usepackage{graphicx}

\usepackage{tikz}
\usepackage{comment}
\usepackage{amsmath,amssymb} 
\usepackage{color}
\usepackage[colorlinks]{hyperref}

\begin{document}
\pagestyle{headings}
\mainmatter
\def\ECCVSubNumber{2}  

\title{Robust Long-Term Object Tracking via Improved Discriminative Model Prediction} 



\newcommand*\samethanks[1][\value{footnote}]{\footnotemark[#1]}
\titlerunning{RLT-DiMP}
\authorrunning{Seokeon Choi, Junhyun Lee, Yunsung Lee and Alexander Hauptmann}
\author{Seokeon Choi\thanks{This work was done while the authors were visiting researchers at CMU.}\inst{1} \and 
Junhyun Lee\samethanks\inst{2} \and 
Yunsung Lee\samethanks\inst{2} \and 
Alexander Hauptmann\inst{3}}
\institute{
KAIST, South Korea \\
\email{seokeon@kaist.ac.kr} \and
Korea University, South Korea \\
\email{\{ljhyun33, swack9751\}@korea.ac.kr} \and
Carnegie Mellon University, USA \\
\email{alex@cs.cmu.edu}
}

%
%
%
\maketitle

\begin{abstract}

We propose an improved discriminative model prediction method for robust long-term tracking based on a pre-trained short-term tracker. The baseline pre-trained short-term tracker is SuperDiMP which combines the bounding-box regressor of PrDiMP with the standard DiMP classifier. Our tracker RLT-DiMP improves SuperDiMP in the following three aspects: (1) Uncertainty reduction using random erasing: To make our model robust, we exploit an agreement from multiple images after erasing random small rectangular areas as a certainty. And then, we correct the tracking state of our model accordingly. (2) Random search with spatio-temporal constraints: we propose a robust random search method with a score penalty applied to prevent the problem of sudden detection at a distance. (3) Background augmentation for more discriminative feature learning: We augment various backgrounds that are not included in the search area to train a more robust model in the background clutter. In experiments on the VOT-LT2020 benchmark dataset, the proposed method achieves comparable performance to the state-of-the-art long-term trackers. The source code is available at: \href{https://github.com/bismex/RLT-DIMP}{https://github.com/bismex/RLT-DIMP}.

\end{abstract}

\section{Introduction}


Visual Object Tracking (VOT), the task of continuously locating an arbitrary target in the first frame of a video, has been drawn attention in both academic and industrial fields over the last decade \cite{smeulders2013visual,VOT2018,VOT2019}. 
It is because VOT can be widely used in real-world applications such as autonomous vehicles \cite{laurense2017path}, robotics \cite{vsuligoj2014object}, and video surveillance systems \cite{ali2016visual}.
With the advance of deep learning techniques, trackers are not only getting better performance but also used at long-term tracking (minute-level) beyond short-term tracking (second-level).

Besides the length of input videos, the clear difference between short-term tracking and long-term tracking is whether the target exists in the field of view, as reflected in standard benchmark datasets \cite{wu2015object,moudgil2018long}.
In general, short-term trackers are designed on the assumption that the target always appears in every single frame, otherwise, the short-term tracker will drift and fail.
Long-term trackers, on the other hand, need to keep track of the object even if it disappears from the field of view in the middle of the frames.
Consequently, the re-detection module, which localizes the target with a confidence score of its absence, is the essential part of long-term trackers.

Because long-term trackers encounter unpredictable abrupt changes during relatively long sequences, the robustness is the most important property of long-term trackers. 
If the long-term tracker misestimates the location of a target because of visual deformation, there is a high risk of incorrect estimation in the following frame.
Previous researches tried to construct robust modules and strategies in various ways such as P-N learning \cite{TLD}, memory model \cite{MMLT}, and dynamic programming \cite{Siam-R-CNN}.
Those methods are focusing on the robustness against visual deformation.

In this work, we focus on the robustness against the re-detection module itself (i.e. reliability of the tracker's prediction) as well as against visual deformation. 
First, we propose a way to reduce the uncertainty of our model and correct the prediction accordingly.
The estimated location of the target, which is robust against the background noise, would not be changed even if we remove a certain small area of the background.
If the estimation is not robust then it would be changed and not reliable even though the confidence score is high.
In view of these characteristics, our model estimates the location of the target from multiple images with randomly erased the small rectangular area. 
Secondly, we utilize spatio-temporal constraints to adjust the confidence score for robust re-detection.
When the target re-appear after occlusion or disappear, the time-space gap should be related in physical.
For instance, if the estimated location of the target that re-appears in a short time is far from the last observation location, we can say that the estimation is unreliable.
To offset the distortion of both time and space and to make a robust estimation as a result, we explicitly adjust the confidence score by penalizing it.
Finally, we perform background augmentation for more discriminative feature learning in the online stage.

Section \ref{related} provides a brief description of existing short- and long-term trackers.
In Section \ref{method}, we explain the details about how our approach can handle the robustness issues of long-term trackers.
The experimental results with analysis and the conclusion are in Section \ref{experiment} and \ref{conclusion}, respectively.


\section{Related Work}
\label{related}

\subsection{Short-term object tracking}

Visual object tracking (VOT) is a task to track an object in a video when the first frame bounding box of the target object is given. A number of deep convolutional neural network (DCNN) based studies have been conducted, such as MDNet \cite{MDNet} and FCNT \cite{wang2015visual}, showing successful results in the VOT Challenge \cite{VOT2015}. Among DCNN-based studies, Siamese architecture \cite{siamfc,siamrpn,Siam-R-CNN,SiamMask,siamrpn++} satisfies end-to-end training capabilities while also showing high efficiency \cite{VOT2018,VOT2019}. DiMP \cite{DiMP} and PrDiMP \cite{PrDiMP}, motivated from the idea of ATOM \cite{ATOM} that solved the limited target estimation problem of the previous studies, are also Siamese architecture-based model that showed performance improvement in the VOT Challenge.

Most tracking models before ATOM \cite{ATOM} were only focused on the development of powerful classifiers. The problem of accurate target state estimation has been overlooked. To this end, ATOM architecture consists of dedicated target estimation and classification components. This target estimation component is trained to predict the overlap between the target object and an estimated bounding box. High-level knowledge is incorporated into the target estimation through extensive offline learning.

Siamese networks \cite{siamfc,siamrpn,Siam-R-CNN,SiamMask,siamrpn++} have received much attention due to their end-to-end training capabilities and high efficiency. However, Siamese approaches are limited in their inability to incorporate information from the background region or previous tracked frames into the model prediction. To deal with this issue, DiMP \cite{DiMP} takes inspiration from the discriminative online learning procedures \cite{ATOM,beyondcf,MDNet}. DiMP tracking architecture consists of two branches: a target classification branch for distinguishing the target from the background, and a bounding box estimation branch for predicting an accurate target box. The target classification branch is derived from two main principles: (i) a discriminative learning loss promoting robustness in the learned target model and (ii) a powerful optimization strategy ensuring rapid convergence. Bounding box estimation branch is utilized from the overlap maximization based architecture introduced in ATOM \cite{ATOM}.

Most tacking models rely on estimating a state-dependent confidence score, but this value lacks a clear probabilistic interpretation, complicating its use. Therefore, in PrDiMP \cite{PrDiMP}, a probabilistic regression formulation is proposed, and it is applied to track the target. PrDiMP network predicts the conditional probability density of the target state given an input image. Their formulation helps the model to be robust from inaccurate annotations and ambiguities in the task.


\subsection{Long-term object tracking}

The target may disappear in the long-term tracking setting, so a re-detection module is essential. In addition, a robust online update method capable of accommodating changes in the visual appearance of the target dramatically affects performance. Therefore, we shortly introduce how long-term trackers overcome significant problems.

The most representative long-term object tracker is TLD \cite{TLD}, which is divided into tracking (T), learning (L), and detection (D). In the tracking part, the tracker predicts the position of the target object based on the median-flow tracker \cite{median-flow}. In the detection part, the detector judges the existence of the target in a cascade manner over the entire area of the image. Assuming that the tracker and detector can fail, the learning module estimates errors based on P-N learning and trains the trackers and detectors more robustly.

Due to its good performance in both accuracy and speed, various object trackers based on Siamese networks have been proposed. One of those methods, MMLT \cite{MMLT}, is designed for long-term tracking to handle visual deformation and target disappearance. In the tracking step, to accommodate changes in the visual appearance, the target position is estimated by the Siamese features obtained from short-term and long-term memory stores inspired by Atkinson-Shiffrin Memory Model (ASMM) \cite{ASMM}. In the re-detection step, the target is detected in the entire image without the dependency of the previous position. In particular, the coarse-to-fine strategy is adopted for improving speed.

Tracking by re-detection paradigm \cite{avidan2004support,babenko2010robust,grabner2006real,siamrpn++} has a long history, but re-detection is challenging due to the existence of distractor objects that are very similar to the template object. Siam R-CNN \cite{Siam-R-CNN}, an adaptation of Faster R-CNN \cite{FasterRCNN} with a Siamese architecture, has two key methods. First, they introduced a hard example mining procedure which trains the re-detector specifically for difficult distractors. Secondly, dynamic programming is used to select the best object in the current time step based on the complete history of all target objects and distractor object tracklets(short object tracks). While being resistant to tracker drift and being able to immediately re-detect object after the disappearance, Siam R-CNN is able to effectively perform long-term object tracking.

\begin{figure}[t!]
\begin{center}
\includegraphics[width=\linewidth]{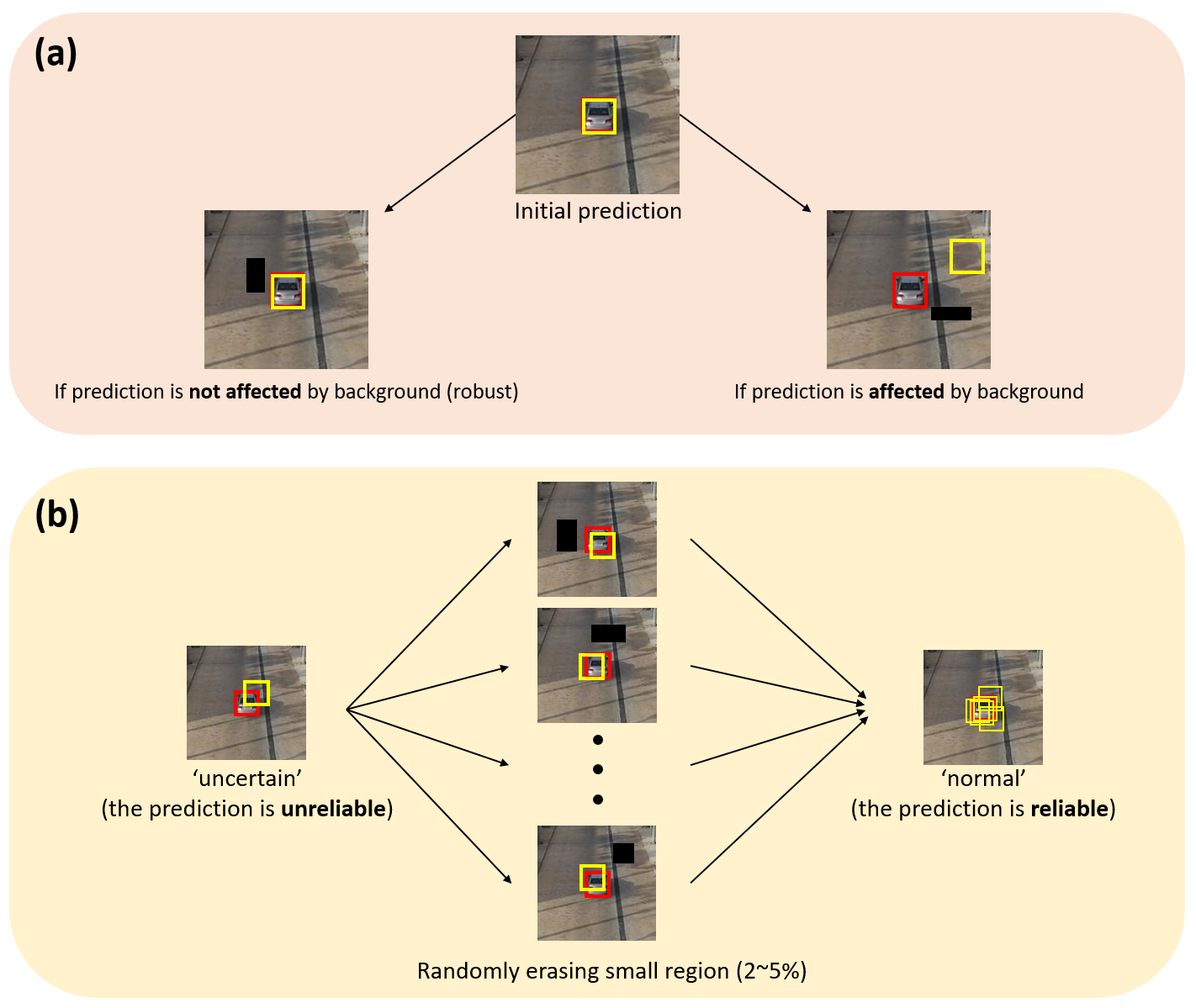}
\end{center}
\vspace{-2.5mm}
\caption{(a) illustrates our assumption that the prediction of the randomly erased image will be changed when the prediction is affected by background features or vice versa. (b) illustrates the concept of how to reduce the uncertainty and correct the prediction.}
\label{fig:uncertainty}
\end{figure}

\section{Proposed Method}
\label{method}
\subsection{Baseline Short-Term Tracker}

We propose an improved Discriminative Model Prediction method for robust long-term tracking based on a pre-trained short-term tracker. Our baseline pre-trained short-term tracker is SuperDiMP \footnote[1]{The pre-trained model is provided at \href{https://github.com/visionml/pytracking}{https://github.com/visionml/pytracking}.} combining the bounding-box regressor of PrDiMP \cite{PrDiMP} with the standard DiMP classifier \cite{DiMP} for better training and inference.

\subsection{Uncertainty Reduction using Random Erasing}

We focus on robustness, which is the consistent generalization (tracking) ability, particularly against the artifact of background features. 
The robust model can consistently track the target whatever occurs in the background, even whenever we remove some region of background. 
We consider uncertainty as an agreement (or consistency) and estimate the location of the target from multiple images after erasing random small rectangular areas.
Because the target usually has a small portion of the whole image, even though we remove a small area, the target might be rarely removed. 
If the prediction of our model changes when the small region of background is randomly deleted, the prediction is affected by the background, so we can say that the certainty is low. 
Adapting this assumption, we correct the flag of our model according to the agreement.

\begin{figure}[t!]
\begin{center}
\includegraphics[width=0.6\linewidth]{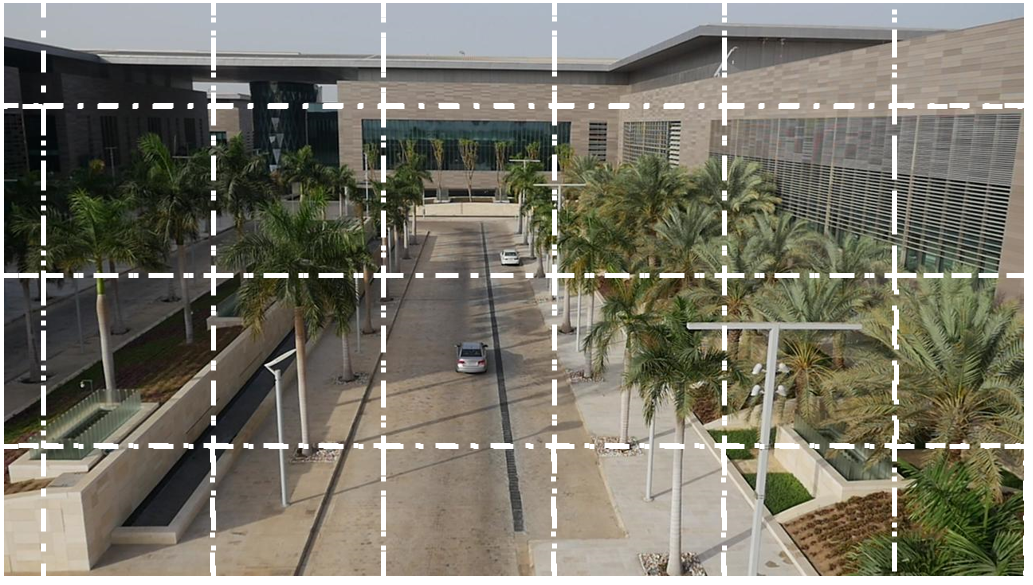}
\end{center}
\vspace{-2.5mm}
\caption{A re-detection example using a general global search method. It takes a long time to find objects in all areas.}
\label{fig:global}
\end{figure}

\subsection{Random Search with Spatio-Temporal Constraints}

\begin{figure}[t!]
\begin{center}
\includegraphics[width=1.0\linewidth]{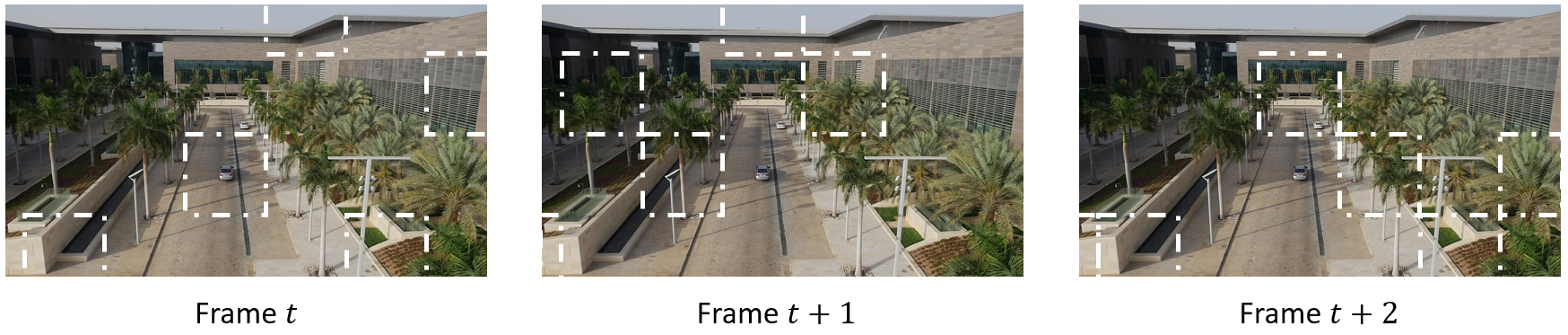}
\end{center}
\caption{Re-detection examples using the proposed random search method. From a stochastic point of view, it is possible to search the entire area within a few frames.}
\label{fig:random}
\end{figure}

\begin{figure}[t]
\begin{center}
\includegraphics[width=1.0\linewidth]{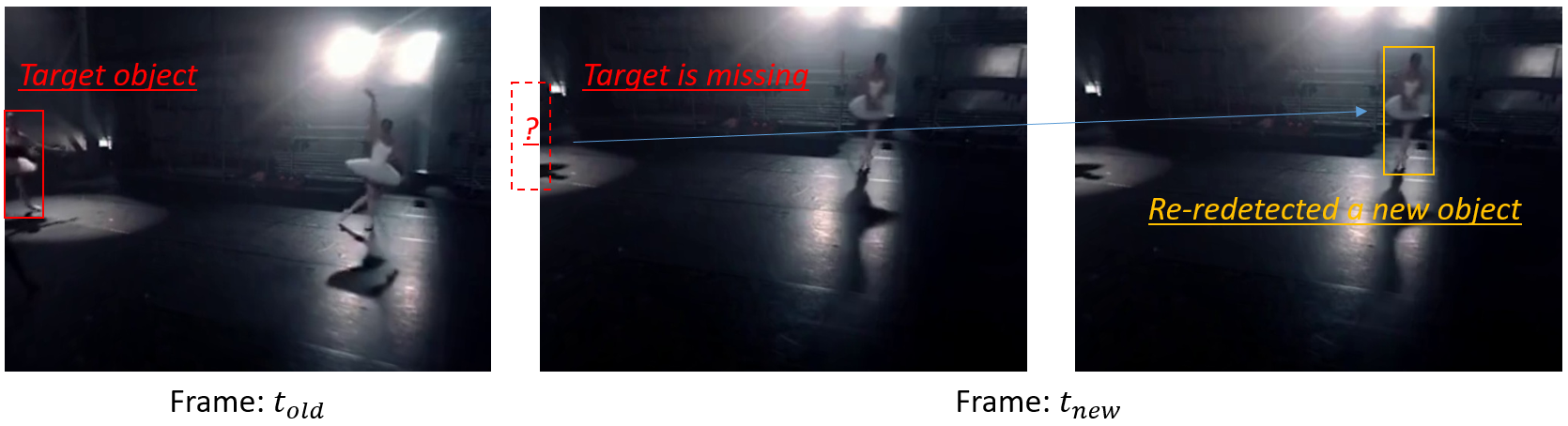}
\end{center}
\caption{Examples of the need for score penalty. In the na\"ive re-detection scheme, a similar object can be re-detected immediately, even if the new object is located far from the original target. In other words, this method is quite vulnerable to background clutter.}
\vspace{-4.5mm}
\label{fig:score}
\end{figure}

General short-term trackers can only find the target if it is included in the searching area. Accordingly, global re-detection capacity is required for robust long-term tracking to deal with occlusions and target disappearances. Most long-term trackers adopt the global sliding window method, as shown in Fig. \ref{fig:global}. However, this approach not only requires high computation costs, but is not robust. To tackle this problem, we propose a robust random search method with spatio-temporal constraints.

\subsubsection{Random search.} 

First, we create global searching templates with a predetermined interval. Next, we adaptively determine the number of searches according to the ratio of the image size to the target size. When the target size is relatively large, we set the fewer number to search. Otherwise, we assign more numbers to search when the target size is relatively small. Then, an object is detected within a randomly selected searching area. As visualized in Fig. \ref{fig:random}, it is possible to search whole regions within a few frames from a stochastic point of view. This stochastic approach improves the re-detection speed compared to general sliding window methods, which is discussed in Section \ref{speed}.

\subsubsection{Score penalty.} 

When the confidence score $s_{new}$ of the newly detected target is higher than a predetermined threshold, the frame in which the new object appeared is designated as the new first frame, and the object is again tracked. Here, we note that this na\"ive re-detection scheme is not robust. Figure \ref{fig:score} shows the examples of sudden detection of other similar objects or background distractors. Once the target is missing, a new object can be re-detected immediately, even if the new object is located far from the original target. However, the probability of an object disappearing and suddenly appearing at a distant location is very low. To prevent this sudden detection, we penalize a confidence score through spatio-temporal constraints, which is expressed as follows:

\begin{equation} 
s'_{new} = w_b (1- w_d \frac{|| \mathbf{p}_{new} - \mathbf{p}_{old} ||_2}{d_{max}} \cdot e^{-w_t | t_{new} - t_{old} |} ) \cdot s_{new},
\label{eq:score}
\end{equation}
\noindent
where $w_b, w_d$, and $w_t$ are hyper-parameters for basic re-detection, distance, and time, respectively. $d_{max}$, $\mathbf{p}$, and $t$ indicate a diagonal length of an image (\emph{i.e.} the maximum distance), a position vector, and a frame number, respectively. In terms of a spatial constraint, the score is more penalized when the distance between new and old positions is large. This distance penalty term prevents the problem of abnormal detection at a long distance. Meanwhile, this distance penalty is compensated by the time penalty (\emph{i.e.} a temporal constraint), as time passes by not finding the target. This is because, if the period during which the object cannot be found becomes longer, the object may newly appear at a longer distance. This temporal constraint allows objects to be detected at relatively far locations. The revised re-detection module makes the tracker more robust.

\subsection{Background Augmentation for More Discriminative Feature Learning}

\begin{figure}[t]
\begin{center}
\includegraphics[width=1.0\linewidth]{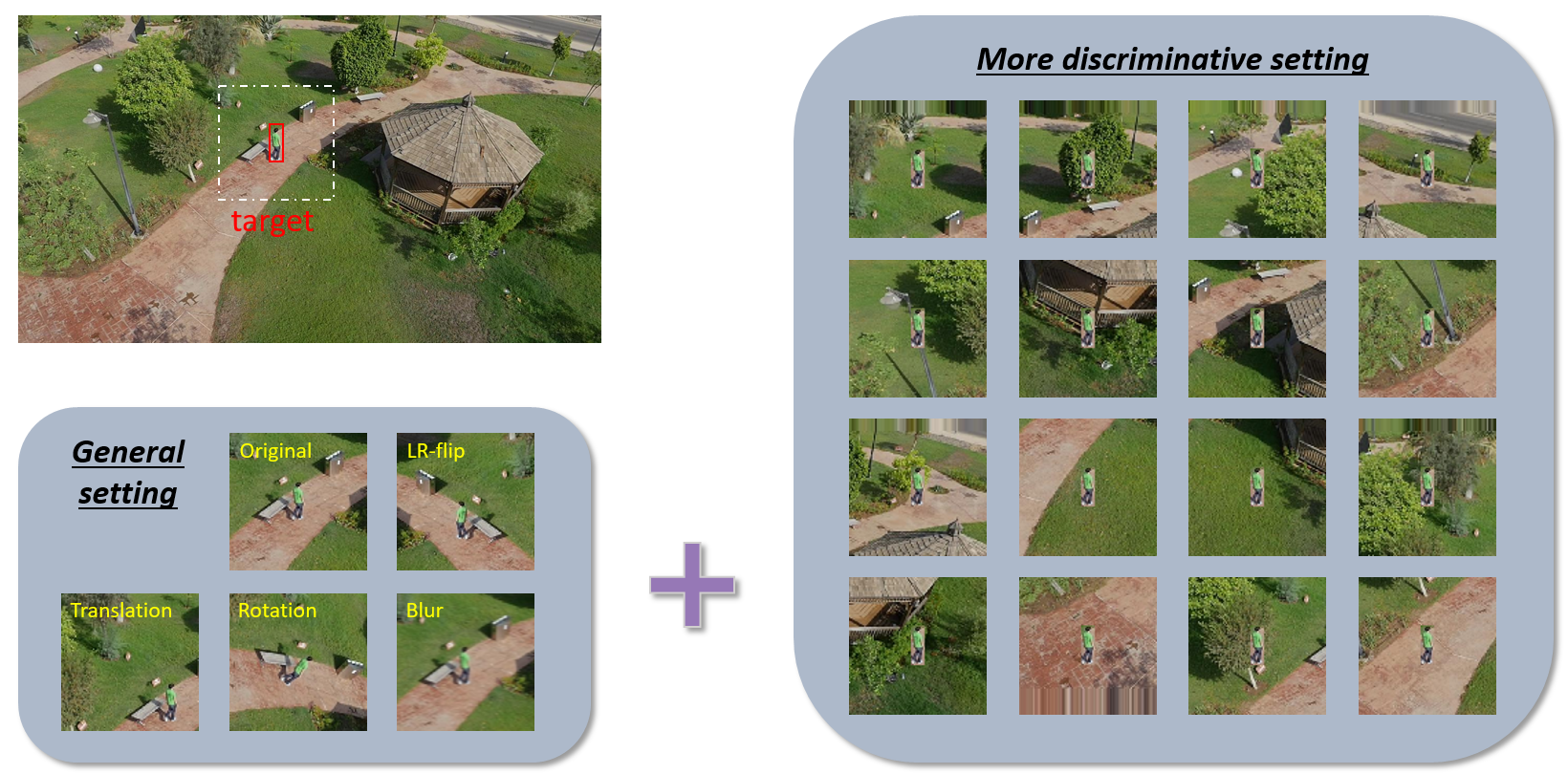}
\end{center}
\vspace{-3.5mm}
\caption{General data augmentation and additional samples for more discriminative learning.}
\vspace{-1.5mm}
\label{fig:discriminative}
\end{figure}

In the short-term baseline tracker, we extract features and trains the model by applying some transforms (blur, rotate, horizontal flip) in the searching area of the target. To train a robust model against the background clutter, we try to augment various backgrounds that are not included in the search area by combining the target image with another background. This augmentation skill makes the model capable of fully exploiting various background appearance information. Figure \ref{fig:discriminative} shows our data augmentation methods for enhanced discriminative feature learning.

\subsubsection{Learning at the first frame.} 

The bounding box is given in the first frame, which means that the target image is completely confident. Therefore, we train the short-term tracker using new data augmentation and additionally store the newly generated images in the memory system. This approach allows for improved discriminative learning in the first frame.

\subsubsection{Online learning.} 

We also perform background augmentation during tracking in the online stage as same as in the first frame. However, this process is only used when the reliability of the bounding box is high since the estimated bounding box is not always confident. The background augmentation process is the same as the first frame one, and the filter is trained if more demanding conditions are met. Unlike in the first frame settings, images with background augmentation are not stored in memory. This prevents the problem of including negative features that cause drifting to other objects.

\subsection{Confidence Score Assignment}

The baseline short-term tracker classifies the state of object tracking into four types according to the score and various conditions: \textit{normal}, \textit{hard negative}, \textit{uncertain}, and \textit{not found}. In the case of \textit{normal} and \textit{hard negative}, the object tracking result is reliable, so the confidence score is given as 1. In the case of \textit{uncertain}, it is difficult to determine whether it is an object or not, so the confidence score is given as 0.5. Lastly, in the case of \textit{not found}, it is predicted that there is no object, so the confidence score is given as 0.

\section{Experiments}
\label{experiment}
\subsection{Dataset and Settings}

\subsubsection{Dataset}

We experiment with a long-term object tracking dataset, LTB50 \cite{LTB}. This dataset is an extension of the LTB35 \cite{LTB} used in the VOT-LT2018 challenge \cite{VOT2018}, and it is officially used in the VOT-LT2019 \cite{VOT2019} challenge. In the VOT-LT2020 challenge, the LTB50 is used unchanged from last year. The LTB50 dataset contains 50 sequences of various objects with a total length of 215,294 frames for single-target object tracking. In each sequence, the target disappears an average of 10 times, and the disappeared target lasts an average of 52 frames. The resolution of video sequences is between 1280 $\times$ 720 and 290 $\times$ 217. All targets are marked with an axis-aligned bounding box. 

\subsubsection{Evaluation protocol}

The proposed RLT-DiMP is evaluated by the evaluation protocol of the VOT-LT2020 benchmark. An evaluation protocol for long-term trackers follows a no-reset protocol, which means that the object tracker will not restart even if the object tracking fails. Three evaluation metrics are adopted for the long-term tracking benchmark: tracking precision, tracking recall, and tracking F-measure. For additional information, please see \cite{LTB}. This evaluation is automatically performed by the VOT toolkit \cite{trax,VOT-toolkit}. All experiments are performed on a system with Intel(R) core(TM) i7-4770 3.40 GHz processor and a single NVIDIA GTX 1080 Ti with 11GB RAM.


\subsection{Quantitative Evaluation}

\begin{table}[t]
\centering
\renewcommand{\arraystretch}{1.15}
\renewcommand{\tabcolsep}{1.0mm}
\begin{tabular}{c|c|ccc}
\hline
Benchmark  & Tracker     & F-score & Precision & Recall \\ \hline
  & FuCoLoT    & 0.411   & 0.507     & 0.346  \\
  & ASINT        & 0.505   & 0.517     & 0.494  \\
  & CooSiam     & 0.508   & 0.482     & 0.537  \\
  & Siamfcos-LT  & 0.520   & 0.493     & 0.549  \\
VOT-LT2019 \cite{VOT2019} & SiamRPNsLT  & 0.556   & 0.749     & 0.443  \\
   & mbdet     & 0.567   & 0.609     & 0.530  \\
   & SiamDW\_LT   & 0.665   & 0.697     & 0.636  \\
   & CLGS        & 0.674   & 0.739     & 0.619  \\
   & LT\_DSE     & 0.695   & 0.715     & 0.677  \\ \hline
VOT-LT2020 & RLT-DiMP   & 0.681   & 0.667     & 0.695  \\ \hline
\end{tabular}
\vspace{2.0mm}
\caption{Comparison with the state-of-the-art methods on VOT-LT2019 and VOT-LT2020 benchmarks. Both benchmarks are based on the LTB50 dataset.}
\vspace{-3.0mm}
\label{table:quantitative}
\end{table}

\begin{figure}[t!]
\begin{center}
\includegraphics[width=1.0\linewidth]{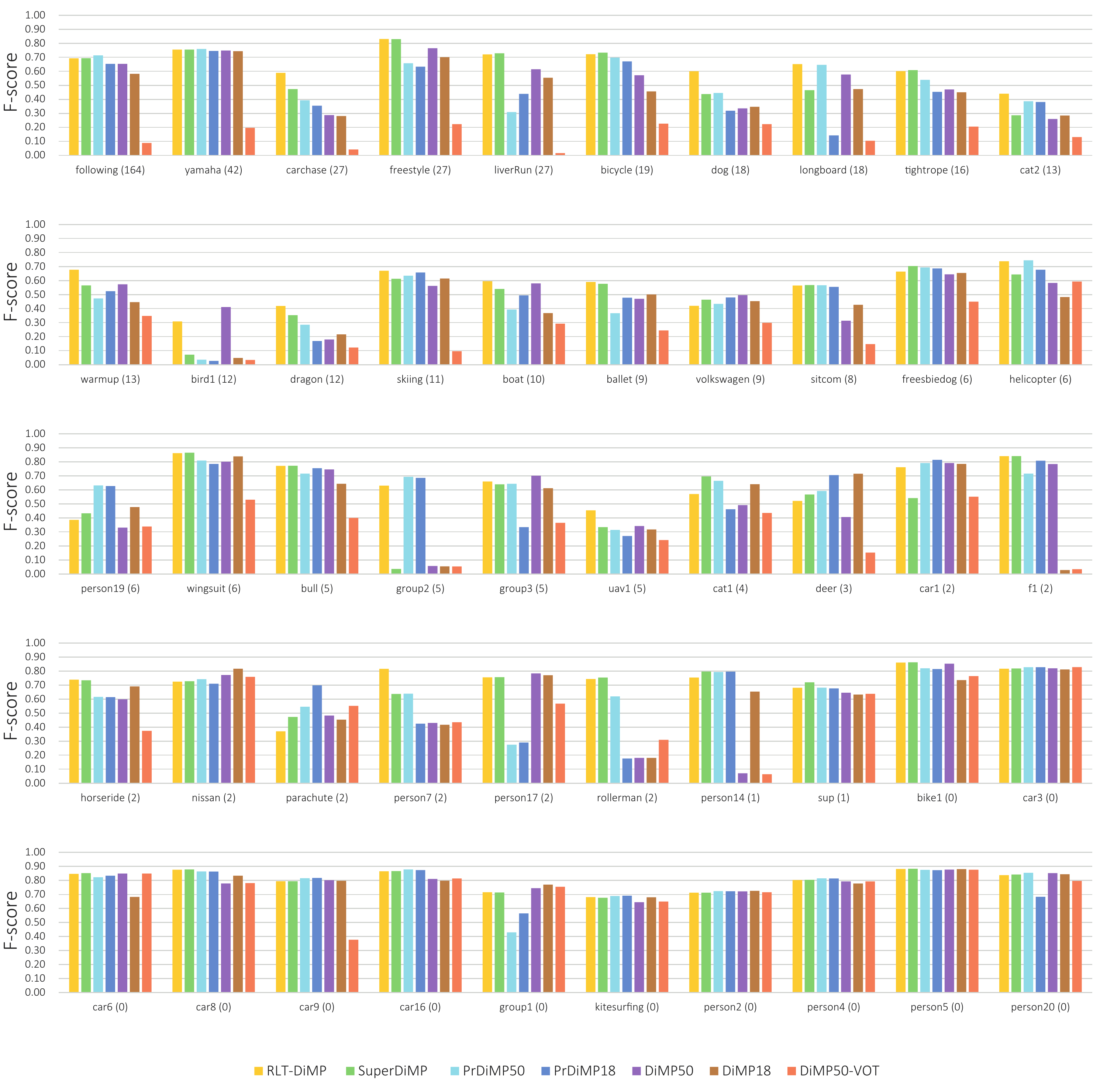}
\end{center}
\vspace{-3.0mm}
\caption{The maximum F-score for each sequence on the LTB50 dataset. Sequences are sorted based on the number of target disappearances, which are indicated by the number in parentheses.}
\label{fig:sequence}
\end{figure}

\subsubsection{Overall comparison with long-term trackers.}

We compare our proposed method with the state-of-the-art methods in the VOT-LT2019 benchmark \cite{VOT2019}. The VOT-LT2019 and VOT-LT2020 benchmarks are based on the LTB50 dataset \cite{LTB}. Both of competing methods and our method are re-detecting long-term trackers (LT1). This term means that all of the trackers detect tracking failure, and an explicit re-detection technique is implemented, unlike the pseudo long-term tracker (LT0). The following taxonomy has been introduced explicitly in \cite{LTB}.

Table \ref{table:quantitative} shows the quantitative evaluation of long-term trackers on the LTB50 dataset. In the VOT-LT2019 benchmark, LT\_DSE tracker has achieved the best F-score and the best tracking recall, and CLGS has achieved the best tracking precision. The tracker LT\_DSE is designed based on target localization by ATOM \cite{ATOM}, bounding box refinement by SiamMask \cite{SiamMask}, and a verifier network by RT-MDNet \cite{RT_MDNET}. The tracker CLGS is designed based on target localization by SiamMask \cite{SiamMask}, global detection by cascade R-CNN \cite{cascade_rcnn}, and an online verifier by RT-MDNet \cite{RT_MDNET}.

Our RLT-DiMP achieves comparable performance to the state-of-the-art long-term trackers in the VOT-LT2019 benchmark. Our method has an F-score of 0.007 higher than that of CLGS and an F-score of 0.014 lower than that of LT\_DSE. For tracking precision, the proposed method achieves lower performance than the top three trackers of the VOT-LT2019 benchmark. Notably, we achieve the highest score in the tracking recall metric, which means that our tracker is well modeled to be robust for long-term tracking through the three contributions: uncertainty reduction using random erasing, robust random search with spatio-temporal constraints, and background augmentation for more discriminative feature learning.

\subsubsection{Comparison by sequence with short-term trackers.}

Our RLT-DiMP method is an extended version based on a pre-trained short-term tracker with improved robustness for long-term object tracking. Our baseline pre-trained short-term tracker is SuperDiMP, which is a combination of the standard DiMP classifier \cite{DiMP} and the bounding-box regressor of PrDiMP \cite{PrDiMP} for better tracking. In this section, we compare our long-term tracker with various short-term trackers, including the baseline tracker SuperDiMP and the individual methods of DiMP \cite{DiMP} and PrDiMP \cite{PrDiMP}. PrDiMP50 and DiMP50 are methods using ResNet50, and PrDiMP18 and DiMP18 are methods using ResNet18 as a backbone. DiMP50-VOT is another version designed to follow a reset protocol for the VOT-ST benchmark. The rest of the PrDiMP- and DiMP-family all follow the no-reset protocol, so these methods can be applied well in a long-term object tracking environment.

Figure \ref{fig:sequence} shows the maximum F-score for each sequence on LTB50. All sequences are listed in order of the number of target disappearances. We can observe that the baseline tracker SuperDiMP has better reasoning skills than the individual methods of DiMP \cite{DiMP} and PrDiMP \cite{PrDiMP}. We emphasize that our RLT-DiMP method outperforms the baseline tracker in almost all sequences. 
This proves that our method is more suitable for long-term object tracking, and robust modeling via uncertainty reduction, robust random search, and background augmentation plays a significant role. This is analyzed in more detail through ablation studies in the next section.

\begin{figure}[t!]
\begin{center}
\includegraphics[width=0.8\linewidth]{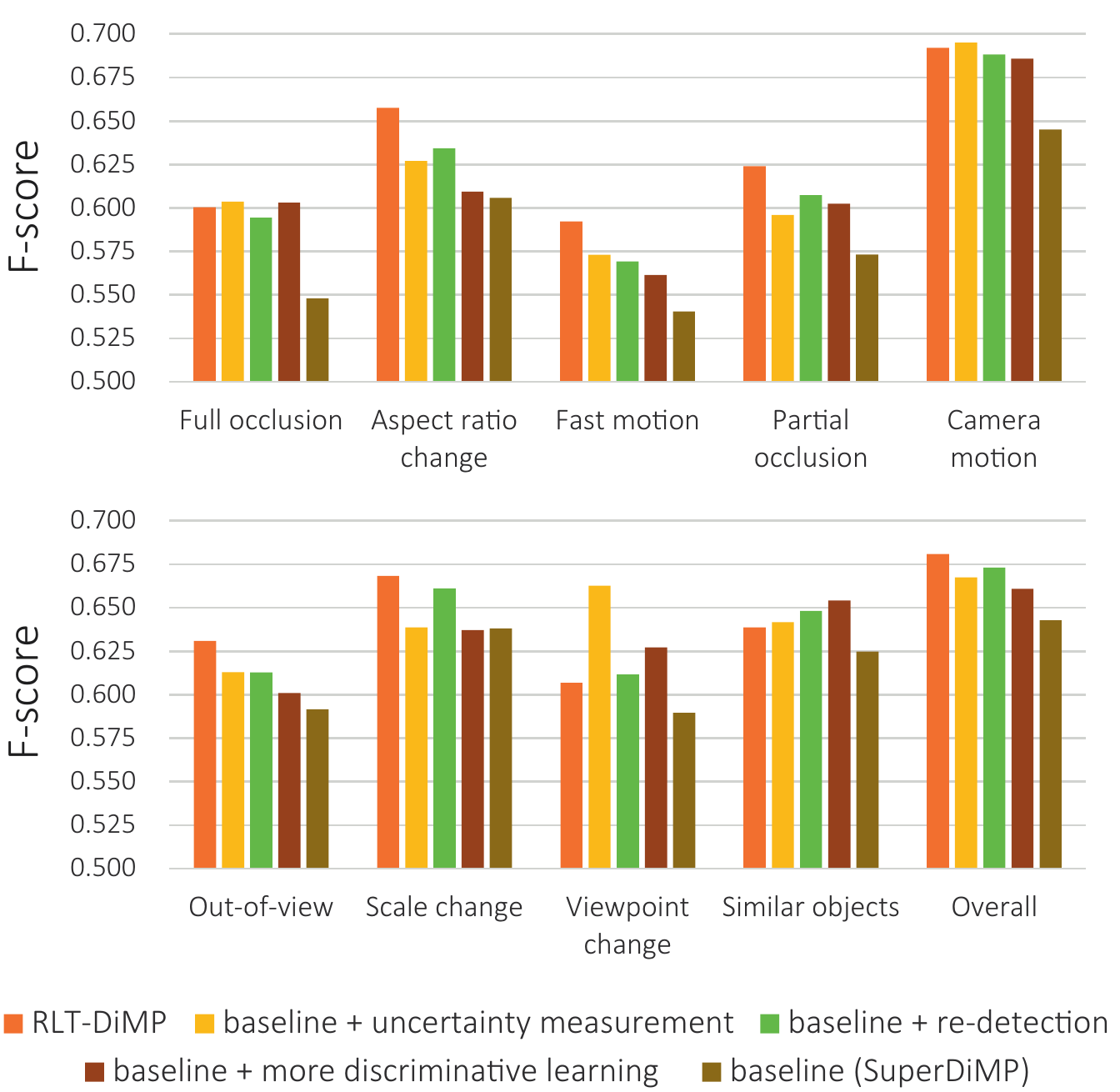}
\end{center}
\vspace{-3.0mm}
\caption{The overall F-score and the average F-score for each attribute on the LTB50 dataset.}
\label{fig:attribute}
\end{figure}

\subsection{Further Evaluations and Analysis}

\subsubsection{Visual attributes on LTB50.}

In the LTB50 dataset, a total of 50 sequences are annotated by nine visual attributes as follows: 1) Full occlusion, 2) Out-of-view, 3) Partial occlusion, 4) Camera motion, 5) Fast motion, 6) Scale change, 7) Aspect ratio change, 8) Viewpoint change, 9) Similar objects. While a \textit{longboard} sequence even has as many as 8 visual attributes, and a \textit{person5} sequence has no visual attribute. As described above, each sequence includes several visual attributes, and we perform an ablation study by averaging performance for each visual characteristic.

\subsubsection{Ablation studies.}

Figure \ref{fig:attribute} shows the overall F-score and the average F-score for each visual attribute with respect to various versions of the proposed method. RLT-DiMP and SuperDimp indicate our final version and baseline version, respectively. In this section, we compare performance by adding one proposed module each from the baseline. 

As described in the overall performance in the bar graph, When applying uncertainty reduction, robust random search, background augmentation algorithms to baseline, the F-score is improved by 0.025, 0.031, and 0.018. Accordingly, these results prove that our sub-algorithms enable our tracker to estimate the target position more robustly. We note that our RLT-DiMP method outperforms the F-score by 0.038 compared to the baseline tracker. 

In the visual attribute analysis, the proposed method surpasses the baseline tracker in all cases. Especially in situations with visual characteristics of full-occlusion, aspect ratio change, fast motion, and partial occlusion, our method has an F-score performance of 0.05 or higher than the baseline.

\subsubsection{Processing time analysis.}
\label{speed}

In the LTB50 dataset, our method records 14.17 fps, which is somewhat lower than the object tracking speed of PrDiMP \cite{PrDiMP} at 21.83 fps and DiMP \cite{DiMP} at 30.22 fps. All three proposed modules inevitably reduce the speed of object tracking. However, in the case of re-detection, we note that the random search method can improve the speed of about 3 fps compared to the global sliding window method.

\subsection{Qualitative Evaluation}

\begin{figure}[t!]
\begin{center}
\includegraphics[width=\linewidth]{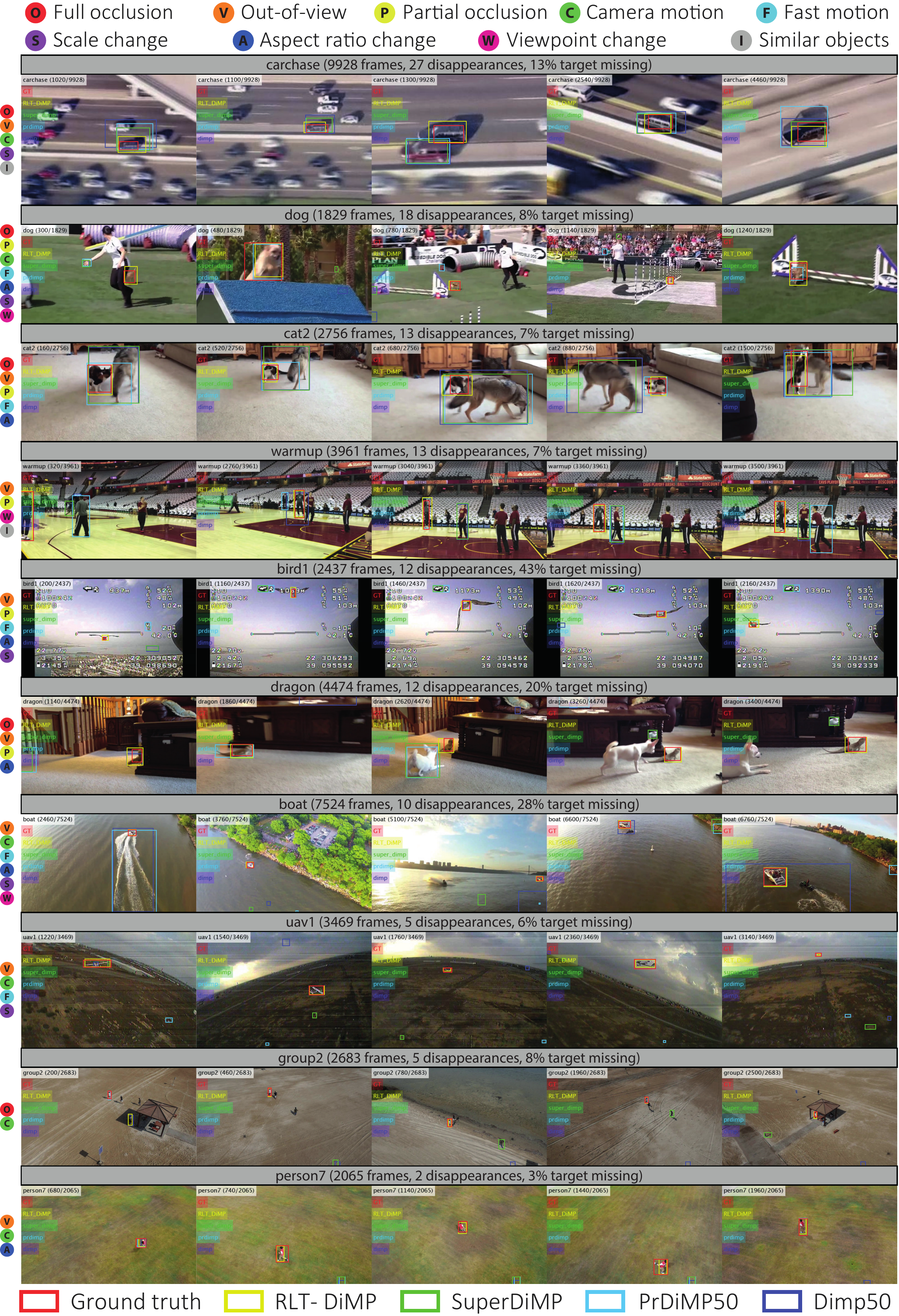}
\end{center}
\vspace{-6.0mm}
\caption{Qualitative results of the proposed RLT-DiMP, SuperDiMP, PrDiMP \cite{PrDiMP}, DiMP \cite{DiMP}. Best viewed in color.}
\label{fig:qualitative}
\end{figure}

In this section, we perform qualitative evaluation by selecting several sequences, which are visualized in Fig \ref{fig:qualitative}. The total of 10 selected sequences is sorted based on the number of times the target disappeared. Besides, not only the number of times disappeared for each sequence, but also the number of frames, the duration of the target disappearance, and the visual attributes are organized in the figure.

We show through various examples that our method RLT-DiMP estimates a target position more accurately than SuperDiMP, PrDiMP \cite{PrDiMP}, and DiMP \cite{DiMP}, without depending on the above situations. In particular, we emphasize that our method has a robust re-detection ability with the aid of spatio-temporal contraints in \textit{dog}, \textit{bird1}, \textit{boat}, and \textit{person7} sequences where the target object suddenly re-appears in a different location. In addition, we note that the target is robustly tracked in \textit{carchase}, \textit{warmup}, \textit{uav1}, and \textit{group2} sequences with similar objects or background clutter through uncertainty reduction using random erasing and enhanced discriminative feature learning with background augmentation.

\section{Conclusion}
\label{conclusion}
We have proposed a robust long-term object tracker via an improved discriminative model prediction method. Conventional object trackers easily follow other objects when the target object disappears from view or is partially obscured due to the presence of background distractors or similar objects. Long-term object trackers need to be robust to these challenging issues because they have to track objects without restarting for a long period. To this end, our approach improves robustness for long-term tracking through uncertainty reduction using random erasing, robust random search with spatio-temporal constraints, and background augmentation for more discriminative feature learning. Quantitative and qualitative evaluation on the VOT-LT2020 benchmark dataset demonstrates the superiority of our method over the state-of-the-art long-term trackers.

\section*{Acknowledgment}

This work was supported in part through NSF grant IIS-1650994, the financial assistance award 60NANB17D156 from U.S. Department of Commerce, National Institute of Standards and Technology (NIST) and by the Intelligence Advanced Research Projects Activity (IARPA) via Department of Interior/Interior Business Center (DOI/IBC) contract number D17PC0034. The U.S. Government is authorized to reproduce and distribute reprints for Governmental purposes notwithstanding any copy-right annotation/herein. Disclaimer: The views and conclusions contained herein are those of the authors and should not be interpreted as representing the official policies or endorsements, either expressed or implied, of NIST, IARPA, NSF, DOI/IBC, or the U.S. Government.

\clearpage
%
%
\bibliographystyle{splncs04}
\bibliography{egbib}
\end{document}